%% file: paper.tex
\documentclass{article}

\input{preamble.tex}

\usepackage[accepted]{icml2019_for_arxiv}
\usepackage{gensymb}

\icmltitlerunning{Multimodal Data Fusion based on the Global Workspace Theory}

\begin{document}

\twocolumn[
\icmltitle{Multimodal Data Fusion based on the Global Workspace Theory}

\icmlsetsymbol{equal}{*}

\begin{icmlauthorlist}
\icmlauthor{Cong Bao}{ucl,emo} 
\icmlauthor{Zafeirios Fountas}{emo,ucl_wchn}
\icmlauthor{Temitayo Olugbade}{ucl}
\icmlauthor{Nadia Bianchi-Berthouze}{ucl}
\end{icmlauthorlist}

\icmlaffiliation{ucl}{Department of Computer Science, University College London, London, United Kingdom}
\icmlaffiliation{emo}{Emotech Labs, London, United Kingdom}
\icmlaffiliation{ucl_wchn}{Wellcome Centre for Human Neuroimaging, Institute of Neurology, University College London, London, United Kingdom}

\icmlcorrespondingauthor{Cong Bao}{cong.bao.18@ucl.ac.uk}
\icmlcorrespondingauthor{Zafeirios Fountas}{f@emotech.co}
\icmlcorrespondingauthor{Temitayo Olugbade}{temitayo.olugbade.13@ucl.ac.uk}
\icmlcorrespondingauthor{Nadia Berthouze}{nadia.berthouze@ucl.ac.uk}

\icmlkeywords{Machine Learning, ICML, Multimodal Fusion, Global Workspace Theory}

\vskip 0.3in
]

\printAffiliationsAndNotice{}

\begin{abstract}

We propose a novel neural network architecture, named the Global Workspace Network (GWN), which addresses the challenge of dynamic and unspecified uncertainties in multimodal data fusion. Our GWN is a model of attention across modalities and evolving through time, and is inspired by the well-established Global Workspace Theory from the field of cognitive science. The GWN achieved average F1 score of 0.92 for discrimination between pain patients and healthy participants and average F1 score = 0.75 for further classification of three pain levels for a patient, both based on the multimodal EmoPain dataset captured from people with chronic pain and healthy people performing different types of exercise movements in unconstrained settings. In these tasks, the GWN significantly outperforms the typical fusion approach of merging by concatenation. We further provide extensive analysis of the behaviour of the GWN and its ability to address uncertainties (hidden noise) in multimodal data.

\end{abstract}

\section{Introduction}
Reasoning about and interpreting multiple sources of information concurrently is an important task in machine learning research as life involves streaming of data from multiple modalities~\cite{Baltrusaitis:2017}. Multimodal data fusion, which leverages the combination of multiple modalities, is a valuable strategy ~\cite{Atrey:2010,Calhoun:2016,Hori:2017,Liu:2018}. Its benefits include complementarity of information, higher prediction performance, and robustness~\cite{Baltrusaitis:2017}. However, multimodal fusion comes with challenges; \cite{Lahat:2015} specifies them under two categories: (1) challenges of multimodal data acquisition, and (2) uncertainties (such as noisy modalities, missing values, conflicting information) in multimodal data. The former type of challenges could be managed with later pre-processing, e.g. resampling to reconcile different temporal resolutions across modalities~\cite{Aung:2016}. However, addressing uncertainties in multimodal data requires specialised design of models that can exploit complementarity or discrepancy across modalities~\cite{Lahat:2015}. While there have been approaches such as \cite{Wu2018} that address the particular problem of missing modalities, fusion of multimodal data with varying types or levels of uncertainty (e.g. noise) which are not known apriori has been less investigated. Findings of the efficacy of automatic learning of weights (e.g. some ``importance'' or ``confidence'' metric) for individual input features ~\cite{Wilderjans:2011,simsekli:2013,Liberman:2014,Kumar:2007,Acar:2011}, the basis of attention mechanisms in machine learning~\cite{bahdanau:2014}, suggests that this may be a more relevant approach to factoring uncertainties into multimodal data fusion. However, while uncertainty also evolves through time ~\cite{Lahat:2015}, the typical attention approach has been uni-dimensional, i.e. attention across modalities alone or attention over time within individual modalities, e.g. in \cite{Beard:2018}. Few studies have explored the propagation of attention across modalities through time. The memory fusion network of \cite{zadeh2018memory} which is based on a cross-modality attention module with a memory is one of such rare cases. 


To address this gap in multimodal data fusion, we propose the Global Workspace Network (GWN) which, like \cite{zadeh2018memory}, propagates cross-modality attention through time. However, unlike previous work, the GWN further addresses the problem of differences in feature dimensionalities of the modalities via a common feature space, based on pre-trained autoencoders. In addition, different from \cite{zadeh2018memory}, our approach is bio-inspired (grounded in the Global Workspace Theory ~\cite{Baars:1997,Baars:2002}) and we implement the GWN's cross-modality attention using the widely-tested transformer architecture~\cite{Vaswani:2017}. 

The Global Workspace Theory (GWT) is a well-developed framework (originally proposed as a model of human consciousness~\cite{Baars:1988}) in cognitive science. The GWT states that concomitant cognitive processes \textit{compete} for the opportunity to \textit{broadcast} their current state (to peer processes)~\cite{Fountas:2011}. At each iteration, the winner (a single process or a coalition of processes) earns the privilege of contributing current information in a \textit{global workspace} which can be accessed by all processes (including the winner)~\cite{Shanahan:2008}. This competition and broadcast cycle is believed to be ubiquitous in the perceptual regions of the brain~\cite{Baars:1988}. Although the literature on GWT includes architectures of biologically-realistic spiking neural networks  \cite{Shanahan:2008,Fountas:2011}, to our knowledge, there has been no direct implementation in machine learning. For such implementation, the GWT can be conceptualised as the combination of a compete-and-broadcast procedure and an external memory structure. In contrast to the global workspace, which can be seen as a communication module, the external memory stores information for later use~\cite{Shanahan:2006}. By considering each modality in multimodal data as analogous to specialised processes in the brain, the similarity between the compete-and-broadcast cycle and typical cross-modality attention mechanism becomes clear. The repetitiveness of the cycle allows the pattern of attention to evolve over time and, given the external memory module, be used in the primary prediction task of the network.

In our implementation of the GWN, the transformer~\cite{Vaswani:2017} was leveraged to simulate the compete-and-broadcast component of the GWT, and the Long Short-Term Memory (LSTM) neural network~\cite{Hochreiter:1997,Gers:1999} as its external memory. There are 3 key elements of transformers that illustrate their advantage and relevance to the current task. First is a self-attention mechanism~\cite{Cheng:2016,Paulus:2017} that we use as the GWN’s compete-and-broadcast procedure, where each modality independently scores all modalities and integrates the data from them based on the resulting weights. A second merit is the transformer's bagging approach, where multiple attention patterns are learnt in parallel, with the advantage of increased robustness. Finally, a third valuable attribute is its memory-based structure~\cite{Weston:2015,Sukhbaatar:2015}. Drawing from traditional applications in Natural Language Processing question answering tasks~\cite{Sukhbaatar:2015,Miller:2016}, this unit further maps the feature vector into query, key, and value spaces to increase the weighting depth and robustness~\cite{Hu:2018}. This additionally enables distributed competition versus broadcasting computations. In essence, the query and key forms can be used for the competition while broadcast is performed on value form, which can have more expressive information that is not valuable for the competition. As for the external memory module, in contrast to the use of a custom two-gated recurrent network in \cite{zadeh2018memory}, we used the well-established LSTM which has two additional gates~\cite{Lipton:2015}. Finally, unlike \cite{zadeh2018memory}, we provide extensive analysis of the behaviour of the GWN in the presence of varying degrees of uncertainties across modalities and over time.

The contribution of this paper is the GWN architecture which we propose as an approach to fusion of sequential data from multiple modalities. We evaluate the architecture on the EmoPain dataset~\cite{Aung:2016}, which consists of motion capture and electromyography (EMG) data collected from patients with chronic lower back pain and healthy control participants while they performed exercise movements. While the EMG has four feature dimensions, the motion capture data comprises 78 dimensions. Further, we provide analysis of the GWN's outputs, demonstrating its effectiveness in handling uncertainty in data.

The paper is organized as follows. We discuss the state of the art in attention-based machine learning in Section~\ref{sec:litrev}. We then describe in Section~\ref{sec:gwn} the proposed GWN architecture that builds on these and present both validation and analysis of the network in Section~\ref{sec:resultsanddiscussion}. Section~\ref{sec:conclusion} concludes the paper.

\section{Related Work}
\label{sec:litrev}
As earlier-stated, there have been different approaches to multimodal fusion. For example, \cite{Lee:2018}~ simply concatenated vectors from individual encoders for each modality. The architecture of \cite{Wu2018}, which was mainly tested on non-sequential inputs, learns both individual encodings as well as a common encoding for the different modalities. For the joint encoding in \cite{Wu2018}, the individual encodings are merged by multiplication. Rather than cover the literature on multimodal data fusion, we refer the reader to \cite{baltruvsaitis2018multimodal} for a comprehensive review and focus our discussion here on attention-based approaches to multimodal data fusion. 


\textbf{Attention over time in multimodal fusion.} In the literature on neural networks for multimodal data, attention  performed on the time axis is usually done separately for each modality, and the resulting context vectors from each modality are then fused as non-temporal features. A representative case of this approach is the Recursive Recurrent Neural Network (RRNN) architecture proposed by~\cite{Beard:2018}. In their work, different modalities (video, audio, and subtitles) extracted from a subtitled audiovisual dataset were divided into segments of uttered sentences and each segment was used an input to the network. For each modality in a segment, a bi-directional LSTM layer was used to extract features. At a given time step, attention computation is performed for each modality separately and the outputs are  concatenated over all modalities together with the current state of a shared memory, which the authors implemented with a Gated Recurrent Unit (GRU) cell~\cite{Cho:2014}. The outcome is then used to update the state of the memory. An advantage of this work is that since each modality was encoded separately, they do not have to follow a common time axis, which allows each modality to optimally exploit its inherent temporal properties. However, as this method cannot account for attention between modalities, different modalities affect the final prediction equally despite the fact that some modalities could be more noisy than others. Thus, the challenge of the dynamics of uncertainty across modalities remains unsolved.

\textbf{Attention across multiple modalities.} Several studies have modelled the relation between modalities in multimodal fusion. The typical approach ~\cite{Wilderjans:2011,simsekli:2013,Liberman:2014} is the use of modality weighting although not particularly based on attention mechanisms ~\cite{bahdanau:2014}. One study that does explicitly use the attention mechanism is the work of~\cite{Hori:2017} on automatic video description. Their approach leverages attention between different modalities using an encoder-decoder architecture~\cite{bahdanau:2014} with separate encoders for each modality and a single decoder. Features of each modality are encoded separately and the decoder weights them to generate a context vector as an output. A similar study~\cite{Caglayan:2016} applies multimodal attention in neural machine translation where images are leveraged in translating the description texts from one language to another. The image and text modalities were first encoded using pre-trained ResNet-50 ~\cite{He:2015} and bi-directional GRU neural networks~\cite{Cho:2014} respectively. Then, attention scores were computed for these encodings. More recently, authors of~\cite{Maharjan:2018} place an attention layer on top of several modality-specific feature encoding layers to model the importance of different modalities in book genre prediction. There are many other works~\cite{Lu:2016,Lejbolle:2018,Lu:2018,Fan:2019} that leverage this technique, i.e. encoding sequential/temporal data for each modality before computing attention weighting and fusing encoded modality-specific features. While it is appropriate for obtaining modality-specific feature representation, it does not allow in-depth quantification of the complex interactions between modalities through time. 

\textbf{Attention across modalities and through time.} As discussed in the introduction, \cite{zadeh2018memory} addresses the limitation of attention over time alone or across modality only by considering both the interaction of multiple modalities and the temporal variations in this interaction. Their architecture is based on separate time encoding of individual modalities. A cross-modality attention is then computed and applied for each time slice. Instead of a single time step per slice, each slice consists of successive time steps $t$ and $t-1$. The weighted multimodal encodings for a given time slice are then fed into a memory module with retain and update gates which are based on neural networks that have the encodings as input. A recurrent update is done using the gate outputs, the previous memory state, and the proposed memory state which is also the output of a neural network computation on the encodings. The findings of \cite{zadeh2018memory} in a set of ablation studies suggest that propagation of attention through time improves prediction performance.
The GWN architecture that we propose makes further advance with implementation of the cross-modality attention module based on the self-attending, multi-head attention transformer architecture \cite{Vaswani:2017}. The GWN additionally addresses the confounding challenge of different feature and/or temporal dimensionalities across the modalities to be fused. While \cite{zadeh2018memory} evaluate their model on data with such characteristic, they do not clarify how their architecture deals with this. In the GWN, we take the approach of learning a common dimensionality across modalities. Based on further controlled experiments, we also contribute analysis of the effect of noise in one of the modalities. 



\section{Global Workspace Network (GWN)}
\label{sec:gwn}

\begin{figure}[t]
    \centering
    \includegraphics[width=0.5\textwidth]{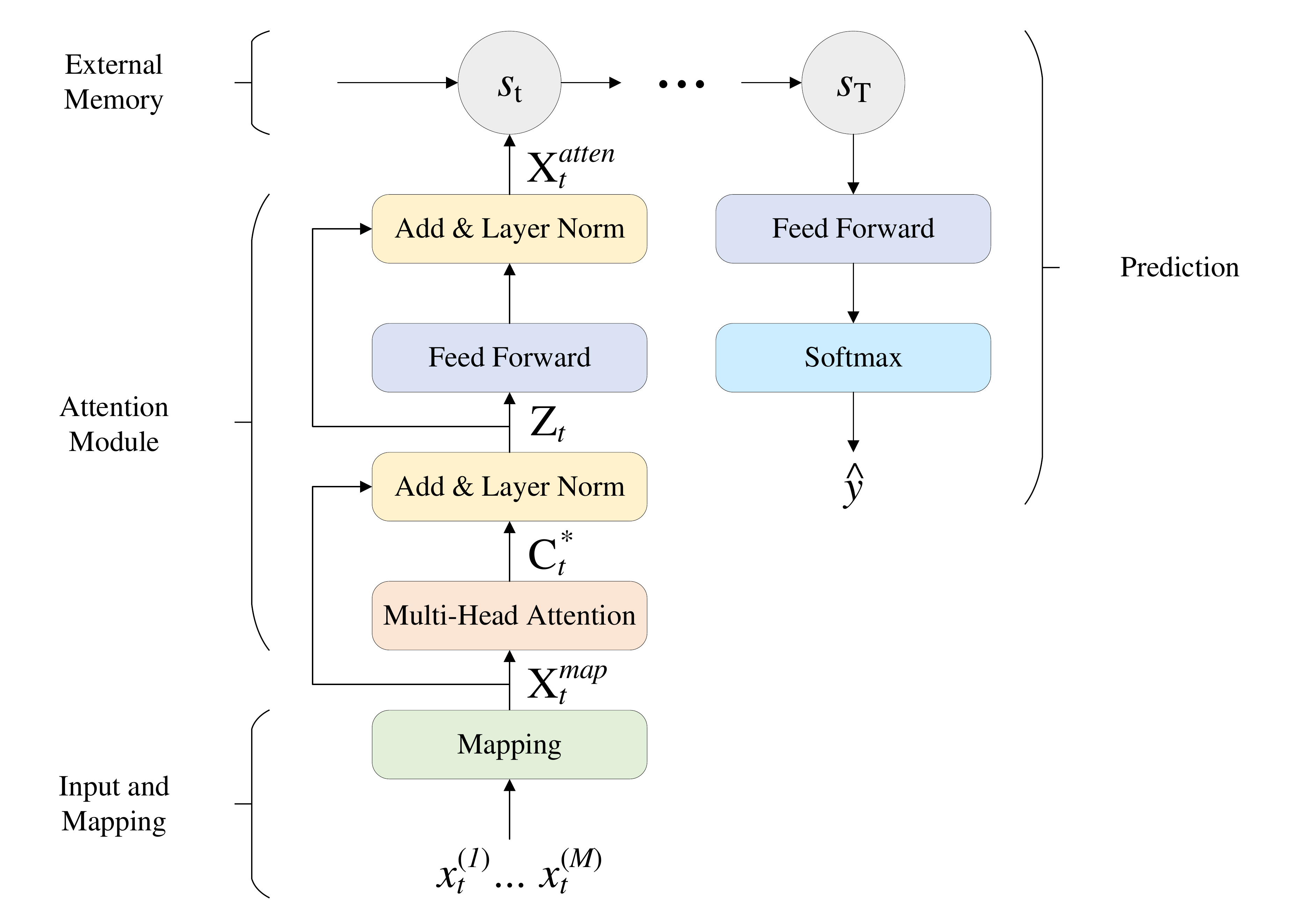}
    \caption{The architecture of the GWN. Here the intermediate matrices $\mat{X}_t^{map}$, $\mat{C_t^*}$, $\mat{Z}_t$, and $\mat{X}_t^{atten}$ have the same dimensionality of $M\times H$.}
    \label{fig:design}
\end{figure}

The architecture of the GWN is shown in Figure~\ref{fig:design}. The network consists of five components: an input unit, a mapping block, an attention module, an external memory, and a prediction block. These components are described in detail below.

\subsection{Mapping Inputs to a Common Feature Space}

Consider $M$ modalities that they have an identical sampling rate, i.e. for each data instance, each modality $m\in M$ in that instance can be written as $\{\Vec{x}_1^{(m)},\dotsc,\Vec{x}_T^{(m)}\}$, where $T$ denotes the common temporal length (common across modalities) of the data instance. The dimensionality at a given time $t$ may nevertheless be different across these modalities, i.e. $\Vec{x}_t^{(m)}\in\bR^{d_m}$. The attention mechanism of the GWN requires identical dimension across modalities and so, it is necessary to have a module for mapping the modalities into the same dimensions.

Inspired by the work of~\cite{Akbari:2018} and~\cite{cong:2018}, we take the approach of using multiple autoencoders~\cite{Vincent:2008} that each learn a common feature space for multiple modalities. Assuming that the common feature space $\Vec{c}$ has a dimensionality of $H$, the mapping function in the encoder for each autoencoder $E^{(m)}$ outputs a vector with dimensionality of $H$. This function can be designed as a feed forward network with one hidden layer which is activated with the rectified linear unit (ReLU)~\cite{Nair:2010} non-linearity, i.e.
\begin{align}
    E^{(m)}\left(\vec{x}_t^{(m)}\right)=\max\left(0,(\vec{x}_t^{(m)}\mat{W}_1+\Vec{b}_1)\right)\mat{W}_2+\Vec{b}_2
\end{align}
where $\vec{x}_t^{(m)}\in\bR^{d_m}$ is the data instance $\vec{x}$ sampled at modality $m$ and time $t$; and $\mat{W}_1$, $\mat{W}_2$, $\Vec{b}_1$, and $\Vec{b}_2$ are trainable parameters of function. The findings of~\cite{Cybenko:1989} suggest that such encoding should be capable of mapping different modalities into a common feature space. $\Vec{c}$ can then be obtained by summing the outputs across the encoders
\begin{align}
    \vec{c}=\sum_mE^{(m)}\left(\vec{x}_t^{(m)}\right)
\end{align}
This is based on previous work in~\cite{cong:2018}. The decoders have the same form as the encoders, i.e.
\begin{align}
    \hat{\vec{x}}_t^{(m)}&=D^{(m)}\left(\vec{x}_t^{(m)}\right) \nonumber \\
    &=\max\left(0,(\vec{c}\mat{W}'_1+\Vec{b}'_1)\right)\mat{W}'_2+\Vec{b}'_2
\end{align}
where $\hat{\vec{x}}_t^{(m)}\in\bR^{d_m}$ is the reconstruction of data instance $\vec{x}$ sampled at modality $m$ and time $t$; and $\mat{W}'_1$, $\mat{W}'_2$, $\Vec{b}'_1$, and $\Vec{b}'_2$ are trainable parameters of decoder. A sum $\cL\left(E^{(m)},D^{(m)}\right)$ of the mean squared error loss for each autoencoder can be used to train the full mapping module.
\begin{align}
    \cL\left(E^{(m)},D^{(m)}\right)=\sum_m\norm{\hat{\vec{x}}_t^{(m)}-\vec{x}_t^{(m)}}^2
\end{align}

\begin{figure}[t]
    \centering
    \includegraphics[width=0.5\textwidth]{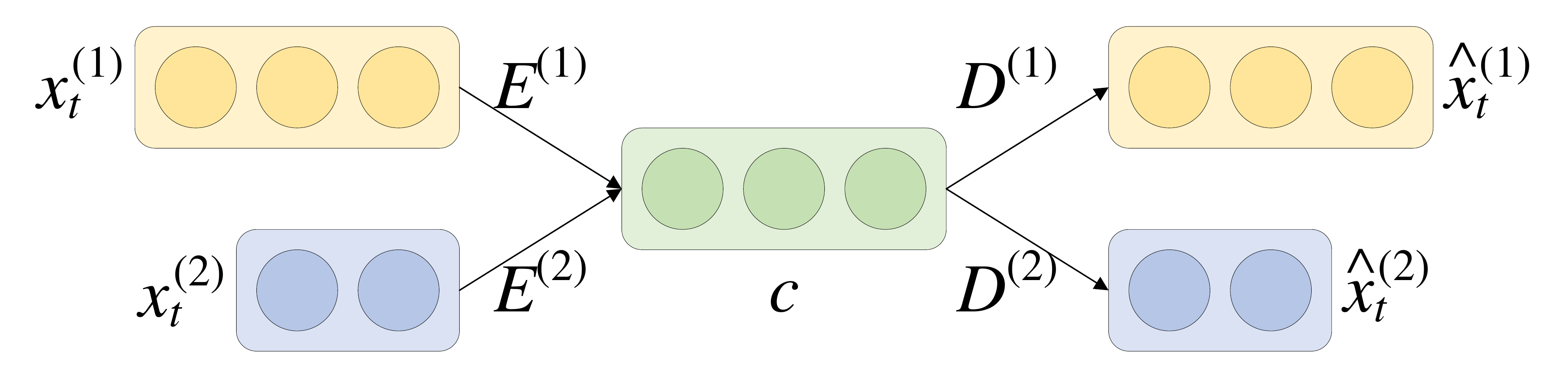}
    \caption{An illustration of the mapping module with two modalities.}
    \label{fig:map}
\end{figure}

Figure~\ref{fig:map} provides an illustration with an example of two modalities mapped into a common feature space and then reconstructed, based on two autoencoders. After pre-training the autoencoders, the encoders are used directly as the mapping function in the GWN. The pre-trained parameters in the encoders then serve as initial values for the mapping block in the GWN. Though this approach introduces more learnable parameters, the findings of~\cite{Hinton:2006} suggest that unsupervised pre-training on shallow layers can improve the performance of a deep network.

For the subsequent attention module, the output vector from each modality's mapping are merged by stacking, to form a matrix $\mat{X}_t^{map}\in\bR^{M\times H}$.

\subsection{The Attention Module}

The attention module is a single layer of the transformer encoder described in ~\cite{Vaswani:2017} with the difference that, in the GWN, the input is a set of different modalities for a number of data instances at a specific time $t$, rather than data sequences (i.e. multiple time steps and instances) based on a single modality. Since the input $\mat{X}_t^{map}\in\bR^{M\times H}$ is already in matrix form, the following multi-head attention calculation can be performed:
\begin{align}
    \mat{C}_t^*=\concat{\mat{C}_t^1,\dotsc,\mat{C}_t^K}\mat{W}^\mathrm{O}
\end{align}
where $K$ is a set of heads and $\mat{W}^\mathrm{O}\in\bR^{KH\times H}$ is a trainable matrix. Each context matrix $\mat{C}_t^k\in\bR^{M\times H}$ for a specific head $k\in K$ is calculated as
\begin{align}
    \mat{C}_t^k=\softmax{\frac{\mat{Q}_t^k\mat{K}_t^{k\mathsf{T}}}{\sqrt{H}}}\mat{V}_t^k \label{eq:appliedattn}
\end{align}
The query, key, and value matrices of a specific head $k$ at time $t$ are calculated as:
\begin{align}
    \mat{Q}_t^k&=\mat{X}_t^{map}\mat{W}_k^\mathrm{Q} \\
    \mat{K}_t^k&=\mat{X}_t^{map}\mat{W}_k^\mathrm{K} \\
    \mat{V}_t^k&=\mat{X}_t^{map}\mat{W}_k^\mathrm{V}
\end{align}
Here, the query, key, and value are variations of the input $\mat{X}_t^{map}$, based on the idea of memory-based attention mechanism~\cite{Miller:2016}. Note that the trainable matrices $\mat{W}_k^\mathrm{Q}\in\bR^{H\times H}$, $\mat{W}_k^\mathrm{K}\in\bR^{H\times H}$, and $\mat{W}_k^\mathrm{V}\in\bR^{H\times H}$ are reused on different time steps $t$ but are independent for different heads $k$.

As shown in Figure~\ref{fig:design}, there are two residual connections~\cite{He:2015} in the attention module. Each of the residual connection is followed by a layer normalisation~\cite{Lei:2016}. The first residual connection can be represented as:
\begin{align}
    \mat{Z}_t=\layernorm{\mat{C}_t^*+\mat{X}_t^{map}}
\end{align}
Here, the assumption of identical dimensionality for residual connection is satisfied as $\mat{C}_t^*\in\bR^{M\times H}$ and $\mat{X}_t^{map}\in\bR^{M\times H}$. The subsequent feed forward layer and the final output of the attention module, respectively, are:
\begin{align}
    \mathrm{FFN}\left(\mat{Z}_t\right)&=\max\left(0,(\mat{Z}_t\mat{W}_1+\vec{b}_1)\right)\mat{W}_2+\vec{b}_2 \\
    \mat{X}_t^{atten}&=\layernorm{\mathrm{FFN}\left(\mat{Z}_t\right)+\mat{Z}_t}
\end{align}
both $\in\bR^{M\times H}$.

\subsection{External Memory}

The external memory is implemented as an LSTM cell~\cite{Hochreiter:1997} with updates:
\begin{align}
    \vec{f}_t&=\sigmoid{[\vec{x}_t^{atten};\vec{h}_{t-1}]\mat{W}^f+\vec{b}^f} \\
    \vec{i}_t&=\sigmoid{[\vec{x}_t^{atten};\vec{h}_{t-1}]\mat{W}^i+\vec{b}^i} \\
    \vec{o}_t&=\sigmoid{[\vec{x}_t^{atten};\vec{h}_{t-1}]\mat{W}^o+\vec{b}^o} \\
    \vec{c}_t&=\vec{f}_t\odot\vec{c}_{t-1}+\vec{i}_t\odot\tanh{[\vec{x}_t^{atten};\vec{h}_{t-1}]\mat{W}^c+\vec{b}^c} \\
    \vec{h}_t&=\vec{o}_t\odot\tanh{\vec{c}_t}
\end{align}
where the input vector $\vec{x}_t^{atten}\in\bR^{MH}$ is the flattened form of $\mat{X}_t^{atten}\in\bR^{M\times H}$; $\sigmoid{\cdot}$, $\tanh{\cdot}$, and $\odot$ are sigmoid, hyperbolic tangent, and Hadamard product (i.e. element-wise product) functions respectively. $\vec{s}_t\in\bR^{2G}$ is the recurrent state at time step $t$, and consists of a memory cell $\vec{c}_t\in\bR^G$ and the output $\vec{h}_t\in\bR^G$ at that time step, with $G$ as an hyperparameter that indicates the size of the external memory. The initial state $\vec{s}_0=[\vec{c}_0;\vec{h}_0]$ is set with zeros. $\vec{f}_t$, $\vec{i}_t$, and $\vec{o}_t$ represent forget, input, and output gates respectively~\cite{Hochreiter:1997,Gers:1999}. All the gates have the same dimensionality $G$. The output vector $\vec{h}_T\in\bR^G$ in the last recurrent state $\vec{s}_T$ is used by the final prediction component.

\subsection{Prediction}

The final prediction module consists of a feed forward layer with one hidden layer activated with a ReLU followed by a softmax function. The layer serves as a simple non-linear transformation from the external memory and can be applied at any time step, making it suitable for online prediction with streaming data. The equations are given as
\begin{align}
    \vec{r}&=\max\left(0,\vec{h}_T\mat{W}_1+\vec{b}_1\right)\mat{W}_2+\vec{b}_2 \\
    \hat{\vec{y}}&=\softmax{\vec{r}}
\end{align}
i.e.
\begin{align}
    \hat{y}_i=\frac{\e{r_i}}{\sum_j\e{r_j}}
\end{align}
where $\vec{r}$ is the prediction result mapped into the distribution $\hat{\vec{y}}$. Both $\vec{r}$ and $\hat{\vec{y}}$ have the same dimensionality, the size of label $L$.

\section{Experiments}
\label{sec:resultsanddiscussion}

To evaluate the proposed GWN architecture, we conducted experiments on the multimodal EmoPain dataset~\cite{Aung:2016}. The dataset, data preprocessing, and experiment tasks are introduced in Section~\ref{sec:4.1}. Section~\ref{sec:4.2} describes the baseline model used for comparison and the methods and metrics of this evaluation. Finally, Section~\ref{sec:4.3} presents the performance and empirical analyses of the GWN.

\subsection{Data}
\label{sec:4.1}

\subsubsection{The EmoPain Dataset}

The EmoPain dataset~\cite{Aung:2016} is suitable for exploring the GWN architecture given that it consists of sequential data from multiple modalities and in unconstrained settings where there are bound to be uncertainties (e.g. in form of sensor noise) in the data, and in varying degrees over time. The data was collected from 22 patients with chronic low back pain and 28 healthy control participants and includes motion capture (MC) and muscle activity data based on surface electromyography (EMG). The data for each participant was acquired while they performed physical exercises that put demands on the lower back. For each exercise, there were two levels of difficulty. There is the normal trial, for 7 types of exercise ((1) balancing on preferred leg, (2) sitting still, (3) reaching forward, (4) standing still, (5) sitting to standing and standing to sitting at preferred pace, (6) bending down, and (7) walking). There is additionally the difficult trial, where four of these exercise types were modified to increase the level of physical demand, i.e. (8) balancing on each leg, (9) holding a 2 kg dumbbell while reaching forward, (10) sitting to standing and return to sitting initiated upon instruction, (11) walking with 2 kg weight in each hand, starting by bending down to pick up the weights, and exercises (2) and (4) repeated without modification. The data was acquired so as to build automatic detection models for pain and related cognitive and affective states, and so after each exercise type, patients self-reported the level of pain they experienced, on a scale of 0 to 10 (0 for no pain and 10 for extreme pain)~\cite{Jensen:1992}. In this paper, we used the subset of the EmoPain dataset with the self-reported pain labels available and where consent was given for further use of the data. This subset consists of 14 patients with chronic pain and 8 healthy control participants, resulting in a total of 200 exercise instances.

\subsubsection{Evaluation Experiment Tasks}

The proposed GWN architecture was evaluated on two classification tasks based on the multimodal EmoPain dataset:

\paragraph{Pain Level Detection Task}

\begin{figure}[t]
    \centering
    \begin{subfigure}[b]{0.225\textwidth}
        \centering
        \hspace*{-1cm}
        \includegraphics[width=\textwidth]{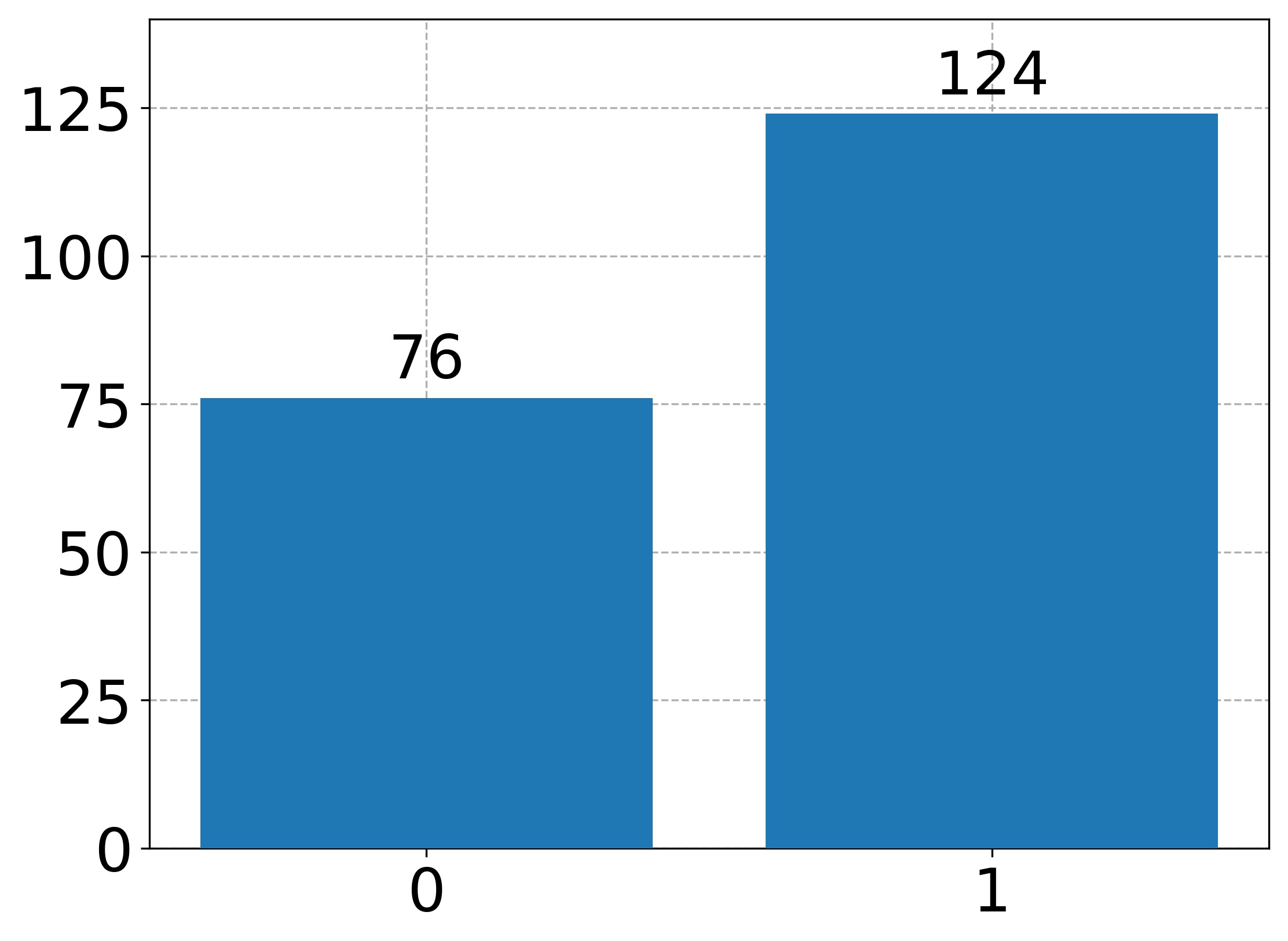}
        \caption{}
        \label{fig:tasks:a}
    \end{subfigure}
    \begin{subfigure}[b]{0.225\textwidth}
        \centering
        \includegraphics[width=\textwidth]{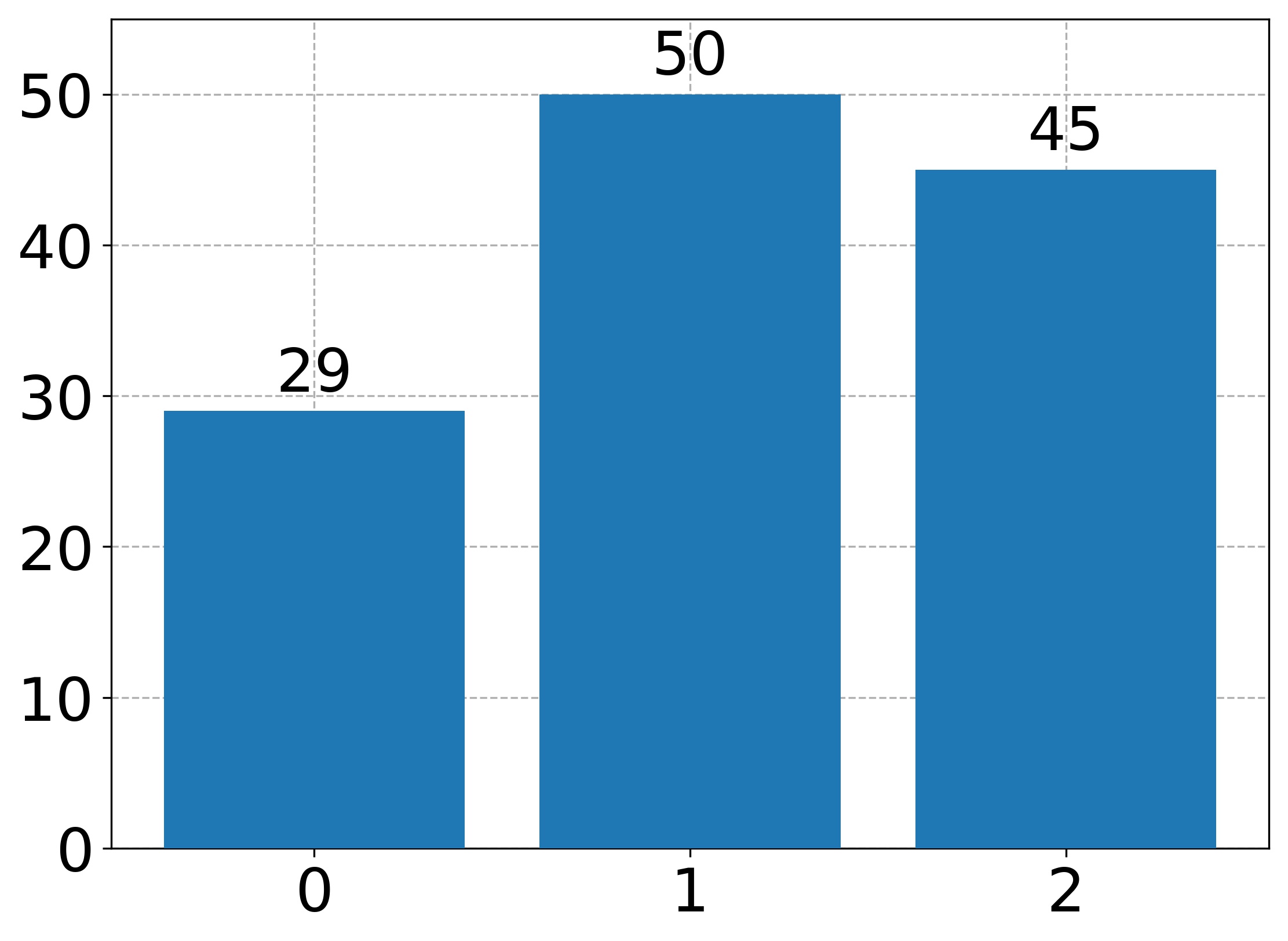}
        \caption{}
        \label{fig:tasks:b}
    \end{subfigure}
    \caption{Number of exercise instances per classes for: (a) The Healthy-vs-Patient Discrimination Task and (b) The Pain Level Detection Task.}
    \label{fig:tasks}
\end{figure}

The aim of this task is to detect the level of a person with chronic pain. The motivation for creating such system is to endow technology with the capability for supporting physical rehabilitation by providing timely feedback or prompts, and personalised recommendations tailored to the pain level of a person with chronic pain. For example, a person with low level pain may be reminded to take breaks at appropriate times and not overdo, whilst a person with high pain may be reminded to breath to reduce tension which may further increase pain levels ~\cite{Olugbade:2019}.

A formal description of the task is as follows. Given M and E, denoting MC and EMG data, for an unseen subject known to have chronic pain (i.e. the event $cp=1$), infer the probability $\p{l|cp=1,M,E}$ that the data corresponds to one of three levels of pain. A random variable $l$ represents the level of chronic pain and is $\in\{0,1,2\}$. In this paper, 0 represents zero level pain, i.e. pain self-report = 0, 1 represents low level pain, i.e 0 $<$ pain self-report $\leq$ 5, and 2
represents high level pain, i.e pain self-report $>$ 5.

\paragraph{Healthy-vs-Patient Discrimination Task}
The healthy control participants were assumed to have no pain. However, patients with chronic pain who reported pain as 0 were not considered to be in the same class as these participants. Hence, a separate model may be needed to first distinguish a person with chronic pain from healthy participants.

The formal definition of the task is as follows. Given M and E, infer the probability $\p{cp|M,E}$ that the data belongs to a person with chronic pain. A random variable $cp$ represents the event that an unseen subject has chronic pain, and $cp\in\{0,1\}$ with 0 for healthy and 1 for  chronic pain person.

Figure~\ref{fig:tasks} shows the number of exercise instances for each class, for the Healthy-vs-Patient Discrimination Task and Pain Level Detection Task respectively.

\begin{table*}[ht]
    \setlength{\tabcolsep}{5.0pt}
    \renewcommand{\arraystretch}{1.5}
    \centering
    \begin{tabular}{c|c|c|c|c|c|c|c|c|c|c}
        \hline
        Task & Validation & Model & ACC & MCC & $\Fone$ (0) & $\Fone$ (1) & $\Fone$ (2) & $\Fone$ (avg) & $r$ & $p$ \\ \hline
        
        \multirow[c]{4}{*}{\makecell{Healthy-vs-Patient \\ Discrimination Task}} &
        \multirow[c]{2}{*}{LOSOCV} &
        CONCATN & 0.765 & 0.489 & 0.662 & 0.820 & - & 0.745 &
        \multirow[c]{2}{*}{0.628} & \multirow[c]{2}{*}{0.003} \\ \cline{3-9}
        & & GWN$^*$ & 0.920 & 0.831 & 0.887 & 0.938 & - & 0.915 & & \\ \cline{2-11}
        & \multirow[c]{2}{*}{$5\times2$ CV} &
        CONCATN & 0.587 & 0.110 & 0.434 & 0.675 & - & 0.555 &
        \multirow[c]{2}{*}{0.768} & \multirow[c]{2}{*}{0.015} \\ \cline{3-9}
        & & GWN$^*$ & 0.648 & 0.225 & 0.482 & 0.733 & - & 0.613 & & \\ \hline
        
        \multirow[c]{4}{*}{\makecell{Pain Level \\ Detection Task}} &
        \multirow[c]{2}{*}{LOSOCV} &
        CONCATN & 0.653 & 0.465 & 0.464 & 0.667 & 0.756 & 0.629 &
        \multirow[c]{2}{*}{0.487} & \multirow[c]{2}{*}{0.068} \\ \cline{3-9}
        & & GWN & 0.766 & 0.645 & 0.581 & 0.800 & 0.857 & 0.748 & & \\ \cline{2-11}
        & \multirow[c]{2}{*}{$5\times2$ CV} &
        CONCATN & 0.395 & 0.075 & 0.249 & 0.438 & 0.441 & 0.379 &
        \multirow[c]{2}{*}{0.596} & \multirow[c]{2}{*}{0.059} \\ \cline{3-9}
        & & GWN$^\dagger$ & 0.448 & 0.151 & 0.309 & 0.474 & 0.503 & 0.430 & & \\ \hline
    \end{tabular}
    \caption{Evaluation experiment results comparing our GWN with the baseline CONCATN. $*$ indicates that a Wilcoxon Signed-Rank test showed that the model performance is  significantly (significance level $p$ = 0.05) higher. $\dagger$ indicates that the model accuracy is marginally significantly higher.}
    \label{tab:results}
\end{table*}

\subsubsection{Data Preprocessing}
Here, we describe the preprocessing performed to prepare the data for the evaluation experiments.

\paragraph{Dealing with A High Sampling Rate}
The EMG data of the EmoPain dataset had been downsampled from 1000Hz to 60Hz for consistency with the MC data. However, 60Hz results in high dimensionality whereas preliminary experiments suggest that 10Hz may be sufficient for the Healthy-vs-Patient Discrimination Task. Thus, we downsampled both MC and EMG data further to 10 Hz to be suitable for the Healthy-vs-Patient Discrimination Task. The original 60Hz was found to be more appropriate for the Pain Level Detection Task.

\paragraph{Padding for Uniform Sequence Lengths}
Based on the findings in~\cite{Dwarampudi:2019,Wang:2019}, we used pre-padding rather than post-padding to obtain uniform time sequence lengths for different data instances. Further, we used zero padding, which is the common approach used in modelling when assuming no prior knowledge about the input data~\cite{Shi:2015}.

\paragraph{Dealing with Imbalanced Data}

As can be seen in Figure~\ref{fig:tasks}, the class distribution of the data is skewed for both pain classification tasks. To reduce bias toward the majority class, we randomly over-sampled data instances of the minority class ~\cite{Kotsiantis:2005}.

\paragraph{Data Augmentation}

The total number of exercise instances available for training and evaluation was 200, which is a limited amount for training a neural network. We employed data augmentation, particularly creating new instances from the original by rotating them, to address this problem. Preliminary experiments that we performed show that rotation about y-axis, which is along the cranial-caudal, outperforms the mirror reflection augmentation used in~\cite{Olugbade:2018}. This augmentation approach used four angles, 0\degree, 90\degree, 180\degree, and 270\degree, and resulted in four times the original data size. For each newly created instance, only the original MC data was changed by the rotation; for these instances, the original EMG data was used unchanged as they are not affected by the orientations.

\subsection{Evaluation Methods}
\label{sec:4.2}

\subsubsection{Baseline Model}

A simple concatenation (CONCATN) architecture, which is representative of the traditional multimodal data fusion approach, was used as the baseline network against which we evaluated our GWN architecture. This baseline allows evaluation of the contribution of the GWN's mapping and attention components to its performance. The CONCATN has identical external memory and  prediction units. Hence, it can be seen as a network that does not pay particular attention to different modalities over time, but rather treats them equally through time.

In the CONCATN, multiple modalities are concatenated along the feature axis and fed into a LSTM network. The feed forward equations are
\begin{align}
    \Vec{x}_t^*&=\concat{\Vec{x}_t^{(1)},\dotsc,\Vec{x}_t^{(M)}} \\
    \Vec{c}_t,\Vec{h}_t&=\lstm{\Vec{x}_t^*,\Vec{c}_{t-1},\Vec{h}_{t-1}}
\end{align}
where $M$ is the number of modalities, $\vec{c}_t$ is a memory cell and $\vec{h}_t$ is the hidden state. Initial states $\vec{c}_0$ and $\vec{h}_0$ have values of zero. Assuming the dimensionality of each modality input at a specific time $t$ is $d_m$, the dimensionality of the concatenated vector $\vec{x}_t^*$ is $\sum_m^Md_m$. The dimensionalities of $\vec{c}_t$ and $\vec{h}_t$ have the same values as in the GWN model. The prediction module is also identical to the GWN model, i.e. the last LSTM output $\vec{h}_T$ is fed into a feed forward network with one hidden layer activated with ReLU~\cite{Nair:2010} non-linearity.

\subsubsection{Validation Technique}

In the experiments carried out, we used the leave-one-subject-out cross-validation (LOSOCV), where the data for a single subject is left out for testing in each fold as is the standard approach for evaluating the generalisation capability of a model to unseen subjects. However, for statistical tests to compare the proposed GWN with the baseline CONCATN, the LOSOCV has the limitation of lack of independence between folds (due to overlapping training sets across folds) that has higher risk of  Type I error~\cite{Dietterich:1998}. Thus, in this work, we additionally perform $5\times 2$ CV (i.e. 5 random replications of 2-fold CV) which has a lower risk of Type I errors~\cite{Dietterich:1998} for the purpose of model comparison. The advantage of the 2-fold CV is that there is no overlap between training sets.

For both LOSOCV and $5\times 2$ CV, we perform Wilcoxon signed-rank test~\cite{Wilcoxon:1945} to compare the proposed GWN and the baseline CONCATN.

\subsection{Results and Discussion}
\label{sec:4.3}

\begin{figure}[t]
    \centering
    \includegraphics[width=0.45\textwidth]{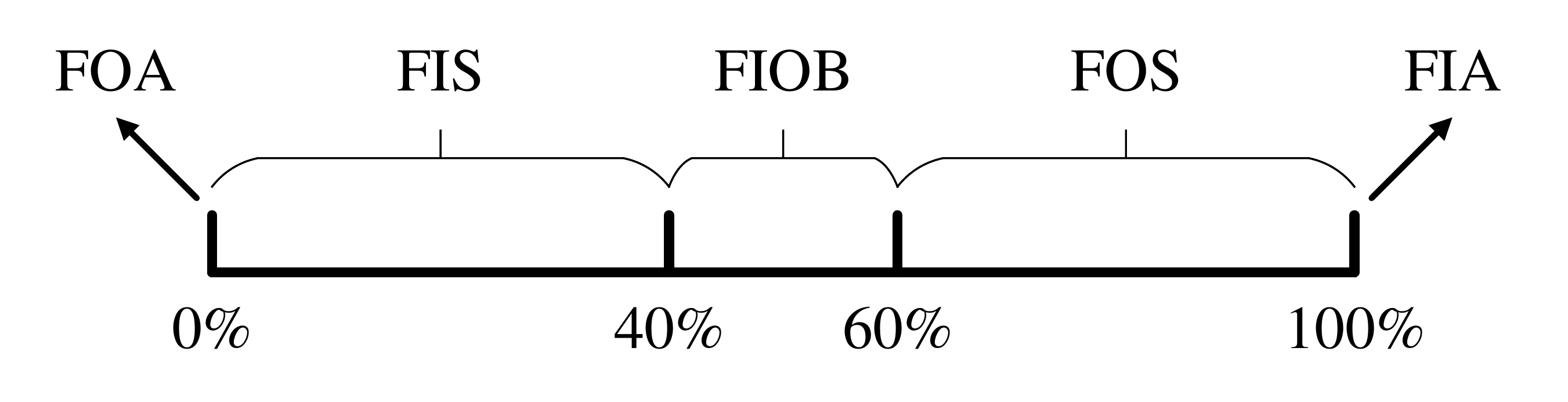}
    \caption{The percentage of itself that a modality pays attention to in the five different attention patterns we found. The thresholds 40\% and 60\% used in this definition were chosen heuristically as a $\pm10\%$ interval around 50\%.}
    \label{fig:cat}
\end{figure}

\begin{figure}[t]
    \centering
    \includegraphics[width=0.45\textwidth]{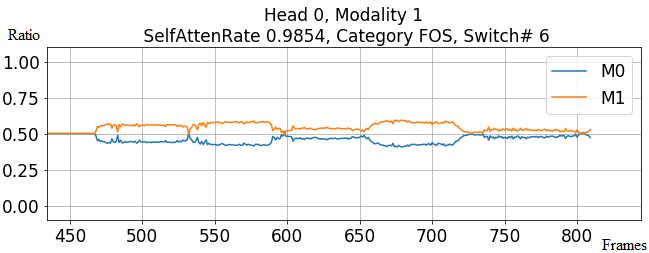}
    \caption{An example of the attention distribution of one exercise instance. Head 0 means the first attention head. Modality 0 (M0) represents MC and modality 1 (M1) represents EMG.}
    \label{fig:cat_example}
\end{figure}

\begin{table*}[t]
    \setlength{\tabcolsep}{5.0pt}
    \renewcommand{\arraystretch}{1.5}
    \centering
    \begin{tabular}{c|c|cc|cc|cc|cc|cc||cc|cc}
        \hline
        1 & \multirow{2}{*}{Noise} & \multicolumn{2}{c|}{FIA} & \multicolumn{2}{c|}{FOS} & \multicolumn{2}{c|}{FIOB} & \multicolumn{2}{c|}{FIS} & \multicolumn{2}{c||}{FOA} & \multicolumn{2}{c|}{\makecell{mean of \\ switch \#}} & \multicolumn{2}{c}{\makecell{std. of \\ switch \#}} \\ \cline{3-16}
        2 & & MC & EMG & MC & EMG & MC & EMG & MC & EMG & MC & EMG & MC & EMG & MC & EMG \\ \cline{2-16}
        3 & None & 0.51 & 0.40 & 0.04 & 0.29 & 0.03 & 0.05 & 0.05 & 0.15 & 0.37 & 0.11 & 0.40 & 14.3 & 1.32 & 30.9 \\ \cline{2-16}
        4 & In MC & 0.31 & 0.43 & 0.08 & 0.36 & 0.02 & 0.05 & 0.11 & 0.10 & 0.48 & 0.07 & 6.92 & 14.6 & 25.0 & 30.4 \\ \cline{2-16}
        5 & In EMG & 0.50 & 0.46 & 0.02 & 0.27 & 0.02 & 0.05 & 0.06 & 0.09 & 0.41 & 0.13 & 0.35 & 12.6 & 1.52 & 30.7 \\ \hline
    \end{tabular}
    \caption{Relative frequency of the five attention patterns for the Pain Level Detection Task, with or without noise added in the data.}
    \label{tab:analysis}
\end{table*}

\subsubsection{Comparison with the Baseline}

Both the GWN and the CONCATN baseline model are trained with Adam optimisation algorithm~\cite{kingma:2014}, learning rate = 0.001, and batch size = 32, which were chosen by grid search. The dimensionality of each LSTM cell are also kept the same, i.e. 64, for the two models. The performance of the GWN can be seen in Table~\ref{tab:results} showing comparison with the CONCATN baseline model, based on accuracy (ACC), Matthews Correlation Coefficient (MCC)~\cite{Matthews:1975}, and F1 scores. 

Our results show that the GWN significantly outperforms the baseline for the Health-vs-Patient Discrimination task (significance level $p$ = 0.05) with F1 score of 0.913 based on LOSOCV, averaged over the two classes. The effect size is $r$=0.768 for the $5\times 2$ CV and $r$=0.628 for the LOSOCV. As expected, due to smaller training data size in the $5\times 2$ CV, it gives lower performance estimation than the LOSOCV for both the baseline CONCATN and the GWN. Although only marginally significant in this case, the GWN also outperforms the baseline CONCATN in the Pain Level Detection Task, effect size $r$=0.596, for the $5\times 2$ CV.

\subsubsection{Attention Patterns}

An additional advantage of the proposed GWN model is that patterns of its attention scores \begin{displaymath}
\mat{a}_t^k=\softmax{\frac{\mat{Q}_t^k\mat{K}_t^{k\mathsf{T}}}{\sqrt{H}}}
\end{displaymath} 
$\forall t$ ($\mat{a}_t^k $ is one of the terms in equation \ref{eq:appliedattn} and $\mat{a}_t^k \in \bR^{M\times M}$) can provide insight into the relevance of each modality through time. In our experiments, we found 5 attention patterns (see Figure~\ref{fig:cat} for further specification of each pattern):

\begin{description}[leftmargin=0cm]
 \item[Favours-Itself-Always (FIA)] The given modality always pays attention to itself and never switches attention to the other modality.
 \item[Favours-Other-Sometimes (FOS)] The given modality mostly pays attention to itself but sometimes switches its attention to the other modality.
 \item[Favours-Itself-and-Other-in-Balance (FIOB)] The given modality pays balanced attention to itself and the other modality.
 \item[Favours-Itself-Sometimes (FIS)] The given modality mostly pays attention to the other modality but sometimes switches attention to itself.
 \item[Favours-Other-Always (FOA)] The given modality always pays attention to the other modality and never to itself.
\end{description}

Figure~\ref{fig:cat_example} gives an example of the FOS pattern. In this case, modality 1 (EMG) pays attention to itself most of the time (98.54\%), with a few switches (6 times) to modality 0 (MC).

The frequency of occurrence of each of the five attention cases are shown in  Table~\ref{tab:analysis} (row 3). It can be seen that MC tends to always pay attention to either only itself or mostly to the EMG (higher FIA and FOA frequencies), whereas the EMG balances its attention (higher FOS, FIOB and FIS frequencies). One possible explanation is that, since the dimensionality of EMG (4) is much lower than the dimensionality of MC data (78), EMG is always trying to balance the difference in information. In contrast, the modality of MC is rich in information, and so can afford to pay 100 percent attention to itself.

\subsubsection{Evaluating How The GWN Deals with Uncertainty in Data}

\begin{table}[t]
    \setlength{\tabcolsep}{3.0pt}
    \renewcommand{\arraystretch}{1.5}
    \centering
    \begin{tabular}{c|c|c|c|c|c|c}
        \hline
        Noise & ACC & MCC & $\Fone$ (0) & $\Fone$ (1) & $\Fone$ (2) & $\Fone$ (avg) \\ \hline
        None & 0.766 & 0.645 & 0.581 & 0.800 & 0.857 & 0.748 \\
        In MC & 0.734 & 0.594 & 0.557 & 0.763 & 0.822 & 0.715 \\
        In EMG & 0.734 & 0.599 & 0.590 & 0.747 & 0.813 & 0.721 \\\hline
    \end{tabular}
    \caption{Results of Pain Level Detection Task with or without noise in each MC and EMG.}
    \label{tab:noise}
\end{table}

In order to further examine the behaviour of the GWN model with respect to uncertainties in the data, noise was added to one modality at a time. We experimented with different levels of noise. We expected that if the GWN manages uncertainty in data, the modality without added noise would pay less attention to the noisy modality. 

The noise was sampled from a Gaussian distribution with zero mean and standard deviation $\sigma_{\text{noise}}$, equal to 10\% of the standard deviation in the original data for this modality. For instance, as the standard deviation of MC in the Pain Level Detection Task is 105.4, in this case, $\sigma_{\text{noise}}=10$ (rounded to the nearest one significant figure number). Similarly, in the case of the EMG recordings of the same dataset, $\sigma_{\text{noise}}=0.001$. 

Table~\ref{tab:noise} presents the result of adding noise. A Wilcoxon Signed-Rank test showed no significant (significance level of $p$ = 0.05) difference between the accuracy of the GWN model with and without noise in the MC data, based on the LOSOCV ($r=0.492,p=0.066$) or with and without noise in the EMG also based on the LOSOCV ($r=0.045,p=0.866$). This suggests that the proposed GWN may be tolerant to this level of noise.

Table~\ref{tab:analysis} shows the GWN's behaviour with the noisy input (row 4 for noisy MC and row 5 for noisy EMG), separated based on the detected attention patterns. Compared with frequencies of the 5 attention cases without added noise, with the noisy MC data, the frequency of FIA for the MC decreases while its frequencies of FOS, FIS, and FOA increase. This indicates that the MC modality is able to recognise noise in itself and rely more on the other modality (EMG). This is also evident in the increase in mean switch frequency.

In contrast, having a noisy EMG (see row 5 in Table~\ref{tab:analysis}) does not result in the same behaviour. Compared with the frequencies of the 5 attention cases (see row 3), the frequency of the EMG's FIA with noisy EMG unexpectedly increases. The frequencies of FOS and FIS also do not increase. Only the FOA frequencies shows expected albeit slight increase. In addition, the mean of switch frequency shows no increment. These results suggest that the EMG modality is less sensitive to its noisiness. One explanation is that the amount of noise added to the EMG data is not sufficient enough to influence the feature representation. Another possible reason is that the system is sensitive to precise amount of information being lost per modality and so since the dimensionalities of MC and EMG are 78 and 4 respectively, the $10\%$ noise added to MC corrupts more information than when added to the EMG, leading to a more sensitive MC in the case of the former.

\section{Conclusion}
\label{sec:conclusion}

We propose the GWN, a novel neural network architecture for multimodal fusion of sequential, multimodal data. Drawing from the Global Workspace Theory, at each time step of the GWN, multiple modalities compete to broadcast information, and each broadcast is propagated through time. We find that this approach outperforms simply concatenating multiple modalities, for pain level detection based on the EmoPain dataset. Our analysis further highlights the modality selectivity that occurs in the GWN for this dataset. Moreover, controlled experiments with simulated noise suggest that the GWN addresses uncertainty and its variation over time. This could be a promising direction for future research in multimodal neural networks while promoting a close connection with cognitive neuroscience research. Such interdisciplinary links may be valuable in consolidating the myriad of advances in both communities.

\section*{Acknowledgements}

The project was partially supported by the Future and Emerging Technologies (FET) Proactive Programme H2020-EU.1.2.2 (Grant agreement 824160; EnTimeMent). It also received support from Emotech Ltd.

\bibliography{paper}
\bibliographystyle{icml2019}

\end{document}

%% file: preamble.tex
\usepackage{amsfonts}
\usepackage{amsmath}
\usepackage{amsopn}
\usepackage{amssymb}
\usepackage{amsthm}
\usepackage{booktabs}
\usepackage{caption}
\usepackage{color}
\usepackage{enumitem}
\usepackage{fancyhdr}
\usepackage{float}
\usepackage{graphicx}
\usepackage{hyperref}
\usepackage{lastpage}
\usepackage{latexsym}
\usepackage{makecell}
\usepackage{microtype}
\usepackage{multirow}
\usepackage{natbib}
\usepackage{setspace}
\usepackage{subcaption}
\usepackage{threeparttable}
\usepackage{tikz}
\usepackage{url}
\usepackage{wrapfig}
\usepackage{xcolor}

\newcommand{\bR}{\mathbb{R}}

\newcommand{\cL}{\mathcal{L}}

\newcommand{\si}{\sigma}

\newcommand{\Fone}{\mathrm{F}_1}

\renewcommand{\emph}[1]{\textit{\textbf{#1}}}

\newcommand{\norm}[1]{\left\vert\left\vert #1 \right\vert\right\vert}

\renewcommand{\tanh}[1]{\mathrm{tanh}\left(#1\right)}

\newcommand{\sigmoid}[1]{\si\left(#1\right)}

\newcommand{\softmax}[1]{\mathrm{softmax}\left(#1\right)}
\newcommand{\concat}[1]{\mathrm{concat}\left(#1\right)}
\newcommand{\layernorm}[1]{\mathrm{layernorm}\left(#1\right)}
\newcommand{\lstm}[1]{\mathrm{lstm}\left(#1\right)}

\newcommand{\p}[1]{p\left(#1\right)}

\newcommand{\e}[1]{\textrm{exp}\left(#1\right)}

\renewcommand{\vec}[1]{\boldsymbol{#1}}
\newcommand{\mat}[1]{\mathbf{#1}}

